\documentclass[journal,transmag]{IEEEtran}
\usepackage{amsmath,amssymb,amsfonts}
\usepackage[ruled,vlined]{algorithm2e}
\usepackage{booktabs}
\usepackage{caption}
\usepackage{capt-of}
\usepackage{comment}
\usepackage{color}
\usepackage{float}
\usepackage{graphicx}
\usepackage[pdftex]{hyperref}
\usepackage{multirow}
\usepackage{nameref}
\usepackage{subfigure}
\usepackage{tabularx}
\usepackage{textcomp}
\usepackage[noadjust]{cite}
\usepackage{algcompatible}

\usepackage{soul}
\usepackage[colorinlistoftodos,prependcaption,textsize=tiny]{todonotes}

\def\BibTeX{{\rm B\kern-.05em{\sc i\kern-.025em b}\kern-.08em
		T\kern-.1667em\lower.7ex\hbox{E}\kern-.125emX}}

\DeclareMathOperator*{\argmin}{arg\,min}

\renewcommand{\COMMENT}[2][.43\linewidth]{%
  \leavevmode\hfill\makebox[#1][l]{//~#2}}
	
\begin{document}
	\title{MetaLabelNet: Learning to Generate Soft-Labels from Noisy-Labels}

	\author{\IEEEauthorblockN{G\"{o}rkem Algan\IEEEauthorrefmark{1},
	Ilkay Ulusoy\IEEEauthorrefmark{1}}
	\IEEEauthorblockA{\IEEEauthorrefmark{1} Middle East Technical University, Electrical and Electronics Engineering, Ankara}
	\thanks{Corresponding author: G. Algan (email: gorkem.algan@metu.edu.tr).}}

	\markboth{IEEE Transactions on Neural Networks and Learning Systems}%
	{Algan: Peerreview paper for IEEE Transactions on Neural Networks and Learning Systems}
	
	\maketitle
	
	\begin{abstract}
		Real-world datasets commonly have noisy labels, which negatively affects the performance of deep neural networks (DNNs). In order to address this problem, we propose a label noise robust learning algorithm, in which the base classifier is trained on soft-labels that are produced according to a meta-objective. In each iteration, before conventional training, the meta-objective reshapes the loss function by changing soft-labels, so that resulting gradient updates would lead to model parameters with minimum loss on meta-data. Soft-labels are generated from extracted features of data instances, and the mapping function is learned by a single layer perceptron (SLP) network, which is called MetaLabelNet. Following, base classifier is trained by using these generated soft-labels. These iterations are repeated for each batch of training data. Our algorithm uses a small amount of clean data as meta-data, which can be obtained effortlessly for many cases. We perform extensive experiments on benchmark datasets with both synthetic and real-world noises. Results show that our approach outperforms existing baselines. 

	\end{abstract}

	\begin{IEEEkeywords}
		deep learning, label noise, noise robust, noise cleansing, meta-learning
	\end{IEEEkeywords}

	\IEEEpeerreviewmaketitle
	\IEEEdisplaynontitleabstractindextext

	\section{Introduction} \label{introduction}
Computer vision systems have made a big leap recently, mostly because of the advancements in deep learning algorithms \cite{krizhevsky2012imagenet,he2016deep,simonyan2014very}. In the presence of large scale data, deep neural networks are considered to have an impressive ability to generalize. Nonetheless, these powerful models are still prone to memorize even complete random noise \cite{zhang2016understanding,krueger2017deep,arpit2017closer}. In the presence of noise, avoiding memorizing this noise and learning representative features becomes an important challenge. There are two types of noises, namely: feature noise and label noise \cite{frenay2014classification}. Generally speaking, label noise is considered to be more harmful than feature noise \cite{zhu2004class}. In this work, we address the issue of training DNNs in the presence of noisy labels.

Various approaches of classification in the presence of noisy labels are summarized in \cite{algan2019image,frenay2014classification}. The most natural solution is to clean noise in the preprocessing stage by human supervision. However, this is not a scalable approach. Furthermore, in some fields, where labeling requires a certain amount of expertise, such as medical imaging, even experts may have contradicting opinions about labels \cite{guan2018said}. Therefore, a commonly used approach is to cleanse the dataset with automated systems \cite{tanaka2018joint,yi2019probabilistic}. In general, instead of noise cleansing in the preprocessing stage, these methods iteratively correct labels during the training. However, these algorithms tackle the problem of differentiating hard informative samples from noisy instances. As a result, even though they provide robustness up to some point, they do not fully utilize the information in the data, which upper-bounds their performance.

In this work, we propose an alternative approach for label correction. Instead of finding and correcting noisy labels, our algorithm aims to generate a new set of soft-labels in order to provide noise-free representation learning. To this end, we propose a reciprocal learning framework, where two networks (MetaLabelNet and the classifier network) are iteratively trained over the feedbacks coming from its peer network. In this context, MetaLabelNet acts like a teacher and learns to produce soft-labels to be used in training for the classifier network. For training MetaLabelNet, a meta-learning paradigm is deployed, which consists of two stages. Firstly, updated classifier parameters are calculated by using soft-labels that are generated by MetaLabelNet. Secondly, the cross-entropy loss on meta-data is calculated by using these updated classifier parameters. Finally, calculated cross-entropy loss on meta-data is backpropagated for the MetaLabelNet. After updating MetaLabelNet, the base classifier is trained by using the soft-labels generated by the MetaLabelNet with the updated parameter set. These steps are repeated consecutively for each batch of data. As a result, MetaLabelNet plays the role of ground truth generator for the base classifier. What differs our approach from the classical teacher-student framework is the usage of meta-learning in the training of MetaLabelNet. Instead of training on the direct feedback coming from the base classifier, we set a meta-objective which deploys two iterations of training in it. As a result, \textit{our meta objective seeks for soft-labels such that gradients from classification loss would lead network parameters in the direction of minimizing meta loss}. Our algorithm differs from conventional label noise cleansing methods in a way that it does not search for clean hard-labels but rather searches for optimal labels in soft-label space to minimize meta-objective. Our contribution to the literature can be summarized as follows:

\begin{itemize}
    \item We propose a novel label noise robust learning algorithm that uses meta-learning paradigm. The proposed meta-objective aims to reshape the loss function by modifying soft-labels, so that learned model parameters are the least noise affected. Our algorithm is model agnostic and can easily be adopted by any gradient-based learning architecture at hand.
    \item To the best of our knowledge, we propose the first algorithm that can work with both noisily labeled and unlabeled data. The proposed framework is highly independent of the training data labels. As a result, it can be applied to unlabeled data in the same manner as labeled data.
    \item We performed extensive experiments on various datasets and various model architectures with synthetic and real-world label noises. Results show the superior performance of our proposed algorithm over state-of-the-art methods.
\end{itemize}

The remainder of the paper is organized as follows. In Section \ref{relatedwork}, we present related work from the literature. The proposed approach is explained in Section \ref{method}, and experimental results are provided in Section \ref{experiments}. Finally, Section \ref{conclusion} concludes the paper by further discussions on the topic.	
	\section{Related Work} \label{relatedwork}
In this chapter, first the related works from the field of deep learning in the presence of noisy labels are presented. Afterward, methods from the meta-learning literature and their usage on noisily labeled data are analyzed. Finally, approaches that make use of both noisily labeled data and unlabeled data are investigated.

\textbf{Learning with noisy labels:} There are various approaches against label noise in the literature \cite{algan2019image}. Some works model the noise as a \textit{probabilistic noise transition matrix} between model predictions and noisy labels \cite{sukhbaatar2014training,xiao2015learning,bekker2016training,misra2016seeing,patrini2017making,xia2019anchor}. However, matrix size increases exponentially with an increasing number of classes, making the problem intractable for large scale datasets. Some researchers focus on regularizing learning by robust loss functions \cite{natarajan2013learning,ghosh2017robust,wang2019improved,zhang2018generalized,wang2019symmetric} or overfit avoidance \cite{ma2018dimensionality}. These methods generally rely on the internal noise robustness of DNNs and aim to improve this robustness further with the proposed loss function. However, they do not consider the fact that DNNs can learn from uninformative random data \cite{zhang2016understanding}. One line of works focuses on increasing the impact of clean samples on training by either picking only clean samples \cite{jiang2017mentornet,han2018progressive,reed2014training,han2018co} or employing a weighting scheme that would up-weight the confidently clean samples \cite{wang2018iterative,lee2018cleannet}. Since these methods focus on low loss samples as reliable data, their learning rate is low and they mostly miss the valuable information from hard samples. One popular approach is to cleanse noisy labels iteratively during training \cite{tanaka2018joint,yi2019probabilistic}. These approaches mainly employ expectation-maximization so that with better classifier better cleansing is provided, and with better labels a better classifier is obtained. Our approach is similar to these approaches in the sense that iterative label correction is employed. However, rather than variations of expectation-maximization, our approach uses meta-learning for label correction. Moreover, we are not interested in finding clean labels but soft-labels that would result in a minimum loss on meta-objective. 

\textbf{Meta learning with noisy labels:} Meta-learning aims to utilize the learning for a meta task, which is a higher-level learning objective than classical supervised learning, e.g., knowledge transfer \cite{hinton2015distilling}, finding optimal weight initialization \cite{finn2017model}. As a meta-learning algorithm, MAML \cite{finn2017model} seeks optimal weight initialization by taking gradient steps on the meta objective. This idea is used to find the most noise-robust weight initialization in \cite{junnan2018learning}. However, their algorithm requires to train ten models simultaneously with two learning loops which is computationally infeasible for most of the time. Three particular works using the MAML approach worth mentioning are \cite{ren2018learning,jenni2018deep,shu2019meta}, in which authors try to find the best sample weighting scheme for noise robustness. Their meta-objective is to minimize the loss on the noise-free meta-data. Therefore, the weighting scheme is determined by the similarity of gradient directions in the noisy training data and clean meta-data. Unlike these methods, our algorithm does not seek for optimal weighting scheme, but rather optimal soft-labels that would provide the most noise robust learning for the base classifier. The logic behind our meta approach is visualized in \autoref{fig:losses}. Similar to our work, \cite{algan2020meta} also uses a meta-objective to update labels progressively. However, they formalize the label posterior with a free variable that does not depend on instances. Differently, we used an SLP network to interpret soft-labels from feature vectors of each instance, which provides a much more stabilized learning, as illustrated in \autoref{fig:mslgvs}. Furthermore, \cite{algan2020meta} can only generate labels for given noisy labels, while our proposed algorithm can generate labels for unseen new data and unlabeled data as well.

\textbf{Learning from both noisily-labeled and unlabeled data:} Some works attempt to use semi-supervised learning techniques on noisily labeled dataset by removing labels of potentially noisy samples \cite{li2020dividemix,nguyen2019self}. Iteratively, unlabeled dataset is updated so that noisy samples are aggregated in the unlabeled dataset and clean samples are aggregated in the labeled dataset. However, these works do not use an additional unlabeled data in combination with noisily labeled data. They rather generate various combinations of unlabeled-subsets from the given labeled data. Differently, during training of MetaLabelNet, our proposed framework does not require training data labels but only meta-data labels. Therefore, the proposed algorithm can be used on unlabeled data in combination with noisily labeled data. To the best of our knowledge, our proposed algorithm is the first one to work with both noisily labeled and unlabeled data.

	\section{The Proposed Method} \label{method}
In this section, we will first give a formal definition of the problem. Afterward, the proposed algorithm is described, and further analyses are provided to support the claim.

\subsection{Problem Statement}
In supervised learning we have a clean dataset $\mathcal{S}=\{(x_1,y_1),...,(x_N,y_N)\}\in (X,Y^h)$ drawn according to an unknown distribution $\mathcal{D}$, over $(X,Y^h)$ where $X$ represents the feature space and $Y^h$ represents the hard-label space for which each label is encoded into one-hot vector. The aim is to find the best mapping function $f:X \rightarrow Y^h$ that is parametrized by $\theta$.
\begin{equation}
    \theta^\star = \underset{\theta}{\argmin}R_{l,\mathcal{D}}(f_\theta)
\end{equation}
where $R_{l,\mathcal{D}}$ is the empirical risk defined for loss function $l$ and distribution $\mathcal{D}$. In the presence of the noise, dataset turns into $\mathcal{S}_n=\{(x_1,\tilde{y}_1),...,(x_N,\tilde{y}_N)\}\in (X,Y^h)$ drawn according to a noisy distribution $\mathcal{D}_n$, over $(X,Y^h)$. Then, risk minimization results in a different parameter set $\theta_n^\star$.
\begin{equation}
    \theta_n^\star = \underset{\theta}{\argmin}R_{l,\mathcal{D}_n}(f_\theta)
\end{equation}
where $R_{l,\mathcal{D}_n}$ is the empirical risk defined for the same loss function $l$ and noisy distribution $\mathcal{D}_n$. Therefore, in the presence of the noise aim is to find $\theta^\star$ while working on noisy distribution $\mathcal{D}_n$. Both $\mathcal{D}$ and $\mathcal{D}_n$ is defined over hard label space $Y^h$. In this work, we are seeking optimal distribution $\hat{\mathcal{D}}$, which is defined over $(X,Y^s)$, where $Y^s$ represents the soft-label space. Since non-zero values ara assigned to each classes, soft-labels can encode more information about the data.

Throughout this paper, we represent given data labels by $y$ and generated soft-labels by $\hat{y}$. While $y$ is in over hard-label space, $\hat{y}$ is in soft-label space. $\mathcal{D}_m$ represents the meta-data distribution. $N$ is the number of training data samples, and $M$ is the number of meta-data samples, where $M<<N$. $N_b$ represents the number of data in each batch and it is same for both training data and meta-data. We represent the number of classes as $C$, and superscript represents the label probability for that class, such that $y_i^j$ represents the label value of $i^{th}$ sample for class $j$.

\begin{figure*}[h]
    \centering
    \resizebox{\textwidth}{!}{\includegraphics[width=\textwidth]{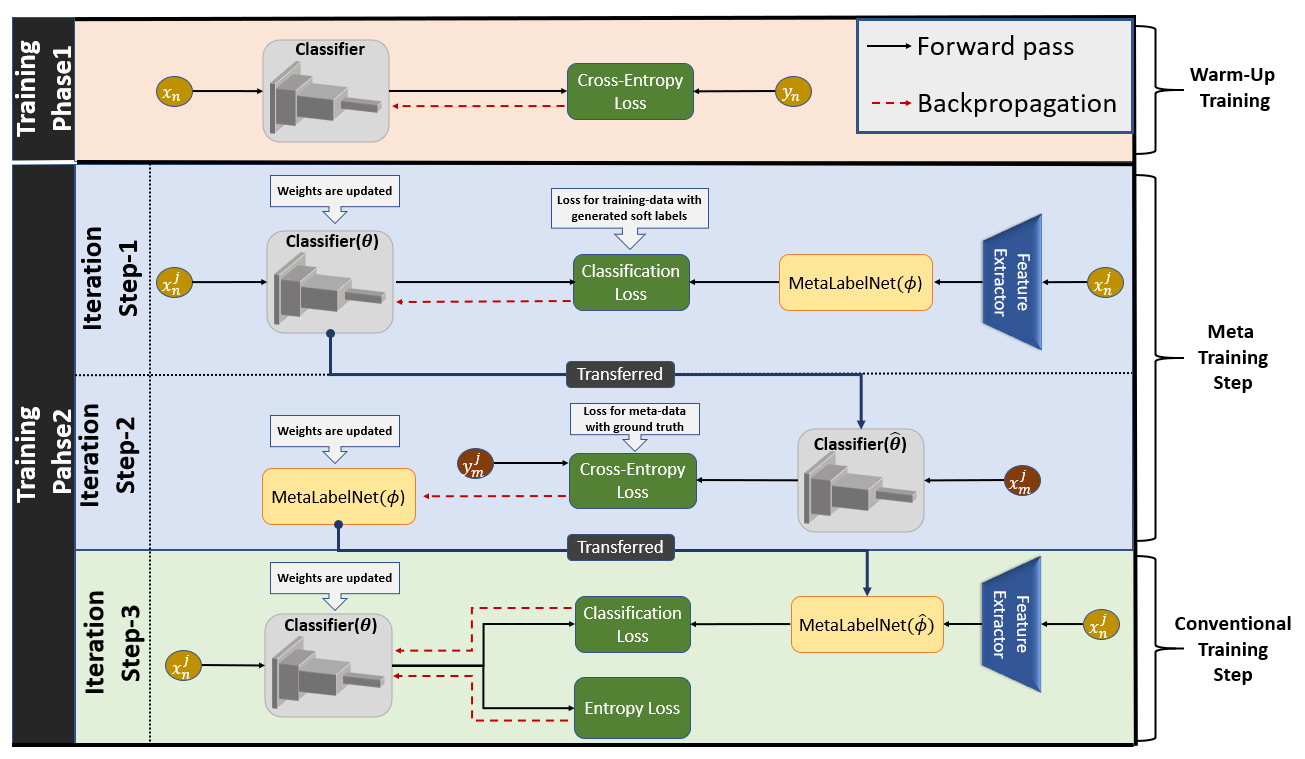}}
    \caption{The overall framework of the proposed algorithm. $x_n, y_n$ represents the training data and label pair. $x^j_n, y^j_n$ represents batches of training data and label pair. $x^j_m, y^j_m$ represents batches of meta-data and label pair. The training consists of two phases. In the first phase (warm-up training), the base classifier is trained on noisy data with the conventional cross-entropy loss. This is useful to provide a stable starting point for the second phase. Training phase-2 consists of 3 consecutive steps, which are repeated for each batch of data. In the first iteration step, classification loss for model predictions is calculated with soft labels produced by MetaLabelNet. Then, updated classifier parameters $\hat{\theta}$ is calculated by SGD. In the second iteration step, the cross-entropy loss for meta-data is calculated using the updated classifier parameters $\hat{\theta}$. Then, updated MetaLabelNet parameters $\hat{\phi}$ is calculated by taking an SGD step on this cross-entropy loss. In the third iteration step, new soft-labels are produced by updated MetaLabelNet parameters $\hat{\phi}$. Then, classifier parameters $\theta$ is updated by the classification loss (with newly generated soft labels) and entropy loss. Feature extractor is the pre-softmax output of the base classifier's duplicate trained at the warm-up phase of the training.}
    \label{fig:overall}
\end{figure*} 

\subsection{Training}
The full pipeline of the proposed framework is illustrated \autoref{fig:overall}. Overall, training consists of two phases. Phase-1 is the warm-up training, which aims to provide a stable initial point for phase-2 to start on. After the first phase, the main algorithm is employed in the second phase that concludes the training. Each of the training phases and their iteration steps are explained in the following.

\subsubsection{Training Phase-1}
It is commonly accepted that, in the presence of noise, deep neural network firstly learn useful representations and then overfit the noise \cite{krueger2017deep,arpit2017closer}. Therefore, before employing the proposed algorithm, we employ a warm-up training for the base classifier on noisy training data with conventional cross-entropy loss. At this stage, we leverage the useful information from the data. This is also beneficial for our meta-training stage. Since we are taking gradients on the feedback coming from the base classifier, without any pre-training, random feedbacks coming from the base network would cause MetaLabelNet to lead in the wrong direction. Furthermore, we need a feature extractor for training in phase-2. For that purpose, the trained network at the end of warm-up training is cloned, and its last layer is removed. Then, this clone network, without any further training, is used as the feature extractor in phase-2.

\subsubsection{Training Phase-2}
This phase is the main training stage of the proposed framework. There are three iteration steps, which are executed on each batch of data consecutively. The first two iterations steps are the meta-training step, in which we update the parameters of MetaLabelNet. Afterward, in the third iteration step, base classifier parameters are updated by using the soft-labels generated with the updated parameters of the MetaLabelNet. These three steps are executed for each batch of data consecutively. The following subsections explain training steps in detail.

\textbf{Meta Training Step:} 
Firstly, data instances are encoded into feature vectors with the help of a feature extractor network $g(.)$.
\begin{equation}
    v = g(x)
    \label{eq:featureextract}
\end{equation}
Then, depending on these encodings, MetaLabelNet $m_\phi(.)$ generates soft-labels.
\begin{equation}
    \hat{y} = m_\phi(v)
    \label{eq:labelpredict}
\end{equation}
Using these generated labels, we calculate posterior model parameters by taking a stochastic gradient descent (SGD) step on the classification loss.
\begin{equation}
        \hat{\theta} = \theta^{(t)} - \nabla_\theta \mathcal{L}_c(f_\theta(x),\hat{y}^{(t)})\Biggr|_{\theta^{(t)}}, \text{and}
    \label{eq:thetahat}
\end{equation}
\begin{equation}
    \mathcal{L}_c(f_\theta(x),\hat{y}^{(t)}) = \dfrac{1}{N_b} \sum_{i=1}^{N_b} l_c(f_\theta(x_i),\hat{y}_i^{(t)})
    \label{eq:classificationloss0}
\end{equation}
where $x_i \in \mathcal{D}_n$ and $\hat{y}_i^{(t)}=m_{\phi^{(t)}}(v_i)$ is the corresponding predicted label at time step $t$. Adapted from \cite{algan2020meta}, we choose the classification loss $l_c$ as KL-divergence loss as follows
\begin{equation}
    l_c =  KL(f_\theta(x_i)||\hat{y}_i), \text{where}
\end{equation}
\begin{equation}
    KL(f_\theta(x_i)||\hat{y}_i) =  \sum_{j=1}^{C} f_\theta^j(x_i) \log \left(\dfrac{f_\theta^j(x_i)}{\hat{y}_i^j}\right)
\end{equation}
Following, we calculate the meta-loss with the feedback coming from updated parameters.
\begin{equation}
    \mathcal{L}_{meta}(f_{\hat{\theta}}(x), y) = \dfrac{1}{N_b} \sum_{j=1}^{N_b} l_{cce} (f_{\hat{\theta}}(x_j), y_j)
    \label{eq:metaloss}
\end{equation}
where $x_j, y_j \in \mathcal{D}_m$ and $l_{cce}$ represents the conventional categorical cross-entropy loss. Finally, we update MetaLabelNet parameters with SGD on the meta loss.
\begin{equation}
    \phi^{(t+1)} = \phi^{(t)} - \beta \nabla_\phi \mathcal{L}_{meta}(f_{\hat{\theta}}(x), y) \Biggr|_{\phi^{(t)}}
    \label{eq:phi}
\end{equation}
Mathematical reasoning for the meta-update is further presented in Section \ref{ap:metaobjective}.

\textbf{Conventional Training Step:} 
In this phase, we train the base network on two losses. The first loss is the classification loss, which ensures that network predictions are consistent with estimated soft-labels.
\begin{equation}
    \mathcal{L}_c(f_\theta(x),\hat{y}^{(t+1)}) = \dfrac{1}{N_b} \sum_{i=1}^{N_b} l_c(f_\theta(x_i),\hat{y}_i^{(t+1)})
    \label{eq:classificationloss}
\end{equation}

Notice that this is the same loss formulation with \autoref{eq:thetahat}, but with updated soft-labels $\hat{y}_i^{(t+1)} = m_{\phi^{(t+1)}}(v_i)$. Moreover, inspired from \cite{tanaka2018joint}, we defined entropy loss as follow.
\begin{equation}
    \mathcal{L}_e(f_\theta(x)) = - \dfrac{1}{N_b} \sum_{i=1}^{N_b} \sum_{j=1}^{C} f^j_\theta(x_i) \log(f^j_\theta(x_i))
    \label{eq:entropyloss}
\end{equation}

Entropy loss forces network predictions to peak only at one class. Since we train the base classifier on predicted soft-labels, which can peak at multiple locations, this is useful to prevent training loop to saturate. Finally, we update base classifier parameters with SGD on these two losses.
\begin{equation}
    \resizebox{0.88\columnwidth}{!}{$
    \theta^{(t+1)} = \theta^{(t)} - \lambda \nabla_\theta \left(\mathcal{L}_{c}(f_\theta(x),\hat{y}^{(t+1)}) + \mathcal{L}_{e}(f_\theta(x))\right)\Biggr|_{\theta^{(t)}}
    $}
    \label{eq:train}
\end{equation}

Notice that we have two separate learning rates $\beta$ and $\lambda$ for MetaLabelNet and the base classifier. Training continues until the given epoch is reached. Since meta-data is only used for the training of MetaLabelNet, it can be used as validation-data for the base classifier. Therefore, we used meta-data for model selection. The overall training for phase-2 is summarized in Algorithm \ref{algo}.

\begin{algorithm*}
    \SetAlgoLined
    \KwIn{Training data $\mathcal{D}_n$, meta-data $\mathcal{D}_m$, batch size $N_b$, phase-1 epoch number $e_{phase1}$, phase-2 epoch number $e_{phase2}$,}
    \KwOut{Base classifier parameters $\theta$, MetaLabelNet parameters $\phi$}
    \BlankLine
    \small
    $epoch = 0$;\\
    \tcp{------------------------------- Phase-1 training -------------------------------}\
    \For(){$epoch< e_{phase1}$}{
        $\theta^{(t+1)} = ConventionalTrain(\theta^{(t)},\mathcal{D}_n)$\COMMENT{Warm-up training with noisy data}\\
        $epoch = epoch + 1$\\
    }
    \BlankLine
    \tcp{------------------------------- Phase-2 training -------------------------------}\
    \For(){$epoch< e_{phase2}$}{
        \ForEach(){batch}{
            $\{x\}$ $\leftarrow$ GetBatch($\mathcal{D}_n$,$N_b$)\\
            $\{x_m,y_m\}$ $\leftarrow$ GetBatch($\mathcal{D}_m$,$N_b$)\\
            \BlankLine
            \tcp{-------------------------- Iteration step-1 ------------------------------}\
            $v = g(x)$\COMMENT{Extract features of data by \autoref{eq:featureextract}}\\
            $\hat{y}^{(t)} = m_\phi(v)$\COMMENT{Generate soft-labels by \autoref{eq:labelpredict}}\\
            $\mathcal{L}_c(f_\theta(x),\hat{y}^{(t)}) = \dfrac{1}{N_b} \sum_{i=1}^{N_b} l_c(f_\theta(x_i),\hat{y}_i^{(t)})$\COMMENT{Classification-loss with soft-labels by \autoref{eq:classificationloss0}}\
            $\hat{\theta} = \theta^{(t)} - \nabla_\theta \mathcal{L}_c(f_\theta(x),\hat{y}^{(t)})$\COMMENT{Updated classifier parameters $\hat{\theta}$ by \autoref{eq:thetahat}}\
            \BlankLine
            \tcp{-------------------------- Iteration step-2 ------------------------------}\
            $\mathcal{L}_{meta}(f_{\hat{\theta}}(x_m), y_m) = \dfrac{1}{N_b} \sum_{j=1}^{N_b} l_{cce} (f_{\hat{\theta}}(x_{m,j}), y_{m,j})$\COMMENT{Meta-loss with $\hat{\theta}$ on meta-data by \autoref{eq:metaloss}}\
            $\phi^{(t+1)} = \phi^{(t)} - \beta \nabla_\phi \mathcal{L}_{meta}(f_{\hat{\theta}}(x_m), y_m)$\COMMENT{Update MetaLabelNet parameters $\phi$ by \autoref{eq:phi}}\
            \BlankLine
            \tcp{-------------------------- Iteration step-3 ------------------------------}\
            $v = g(x)$\COMMENT{Extract features of data by \autoref{eq:featureextract}}\\
            $\hat{y}^{(t)} = m_\phi(v)$\COMMENT{Generate soft-labels by \autoref{eq:labelpredict}}\\
            $\mathcal{L}_c(f_\theta(x),\hat{y}^{(t+1)}) = \dfrac{1}{N_b} \sum_{i=1}^{N_b} l_c(f_\theta(x_i),\hat{y}_i^{(t+1)})$\COMMENT{Classification-loss with new soft-labels by \autoref{eq:classificationloss}}\
            $\mathcal{L}_e(f_\theta(x)) = - \dfrac{1}{N_b} \sum_{i=1}^{N_b} \sum_{j=1}^{C} f^j_\theta(x_i) \log(f^j_\theta(x_i))$\COMMENT{Entropy-loss by \autoref{eq:entropyloss}}\\
            $\theta^{(t+1)} = \theta^{(t)} - \lambda \nabla_\theta \left(\mathcal{L}_{c}(f_\theta(x),\hat{y}^{(t+1)}) + \mathcal{L}_{e}(f_\theta(x))\right)$\COMMENT{Update $\theta$ by \autoref{eq:train}}\\
        }
        $epoch = epoch + 1$;\\
     }
     \caption{Learning with MetaLabelNet}
     \label{algo}
\end{algorithm*}

\subsection{Learning with Unlabeled Data}
The presented algorithm in \autoref{fig:overall} uses training data labels $y_n$ only in the training phase-1 (warm-up training). The training phase-2 is totally independent of the training data labels. Therefore, training phase-2 can be used on the unlabeled data in the same way as the labeled data. As a result, if there exists unlabeled data, warm-up training is conducted on the labeled data. Afterwards, the proposed Algorithm \ref{algo} is applied to both labeled and unlabeled data in the same manner. 

\subsection{Reasoning of Meta-Objective} \label{ap:metaobjective}
We can rewrite the update term for MetaLabelNet as follow.
\begin{equation}
    - \beta \nabla_\phi \mathcal{L}_{meta}(f_{\hat{\theta}}(x), y) \Biggr|_{\phi^{(t)}}
\end{equation}
\begin{equation}
    = - \beta \frac{\partial \mathcal{L}_{meta}(f_{\hat{\theta}}(x), y)}{\partial \hat{\theta}} \frac{\partial \hat{\theta}}{\partial \phi} \Biggr|_{\phi^{(t)}}
\end{equation}
\begin{equation}
    \resizebox{0.88\columnwidth}{!}{$
    = - \beta \frac{\partial \mathcal{L}_{meta}(f_{\hat{\theta}}(x), y)}{\partial \hat{\theta}} \frac{\partial }{\partial \phi} \left(- \frac{\partial \mathcal{L}_c(f_\theta(x),\hat{y}^{(t)})}{\partial \theta} \right) \Biggr|_{\phi^{(t)}}
    $}
\end{equation}

where $\hat{y}^{(t)} = m_{\phi^{(t)}}(g(x))$. $g(.)$ is a fixed encoder for which no learning occurs. All derivatives of $\phi$ are taken at time step $t$. Therefore, we drop the notation $\Biggr|_{\phi^{(t)}}$ from now on.
\begin{equation}
    \resizebox{0.88\columnwidth}{!}{$
    = - \dfrac{\beta}{N_b} \sum_{j=1}^{N_b} \frac{\partial l_{cce}(f_{\hat{\theta}}(x_j), y_j)}{\partial \hat{\theta}} \frac{\partial }{\partial \phi} \left(-\dfrac{1}{N_b} \sum_{i=1}^{N_b} \frac{\partial l_c(f_\theta(x_i),\hat{y}_i^{(t)})}{\partial \theta} \right)$
    }
\end{equation}
\begin{equation}
    \resizebox{0.88\columnwidth}{!}{$
    = \frac{\beta}{N_bN_b} \frac{\partial }{\partial \phi} \left( \sum_{j=1}^{N_b} \frac{\partial l_{cce}(f_{\hat{\theta}}(x_j), y_j)}{\partial \hat{\theta}} \sum_{i=1}^{N_b} \frac{\partial l_c(f_\theta(x_i),\hat{y}_i^{(t)})}{\partial \theta} \right) 
    $}
\end{equation}

\begin{equation}
    \resizebox{0.88\columnwidth}{!}{$
    = \frac{\beta}{N_bN_b} \frac{\partial }{\partial \phi}\left( \sum_{i=1}^{N_b} \sum_{j=1}^{N_b} \frac{\partial l_c(f_\theta(x_i),\hat{y}_i^{(t)})}{\partial \theta} \frac{\partial l_{cce}(f_{\hat{\theta}}(x_j), y_j)}{\partial \hat{\theta}} \right) 
    $}
\end{equation}

Let
\begin{equation}
    \resizebox{0.80\columnwidth}{!}{$
    S_{ij}(\theta,\hat{\theta},\phi^{(t)}) = \frac{\partial l_c(f_\theta(x_i),\hat{y}_i^{(t)})}{\partial \theta} \frac{\partial l_{cce}(f_{\hat{\theta}}(x_j), y_j)}{\partial \hat{\theta}}
    $}
\end{equation}

Then we can rewrite meta update as
\begin{equation}
    \resizebox{0.88\columnwidth}{!}{$
    \phi^{(t+1)} = \phi^{(t)} + \frac{\beta}{N_b} \frac{\partial }{\partial \phi}\left( \sum_{i=1}^{N_b} \frac{1}{N_b} \sum_{j=1}^{N_b} S_{ij}(\theta,\hat{\theta},\phi^{(t)}) \right) 
    $}
    \label{eq:metaobjective}
\end{equation}

\begin{figure}[]
    \centering
    \includegraphics[width=\columnwidth]{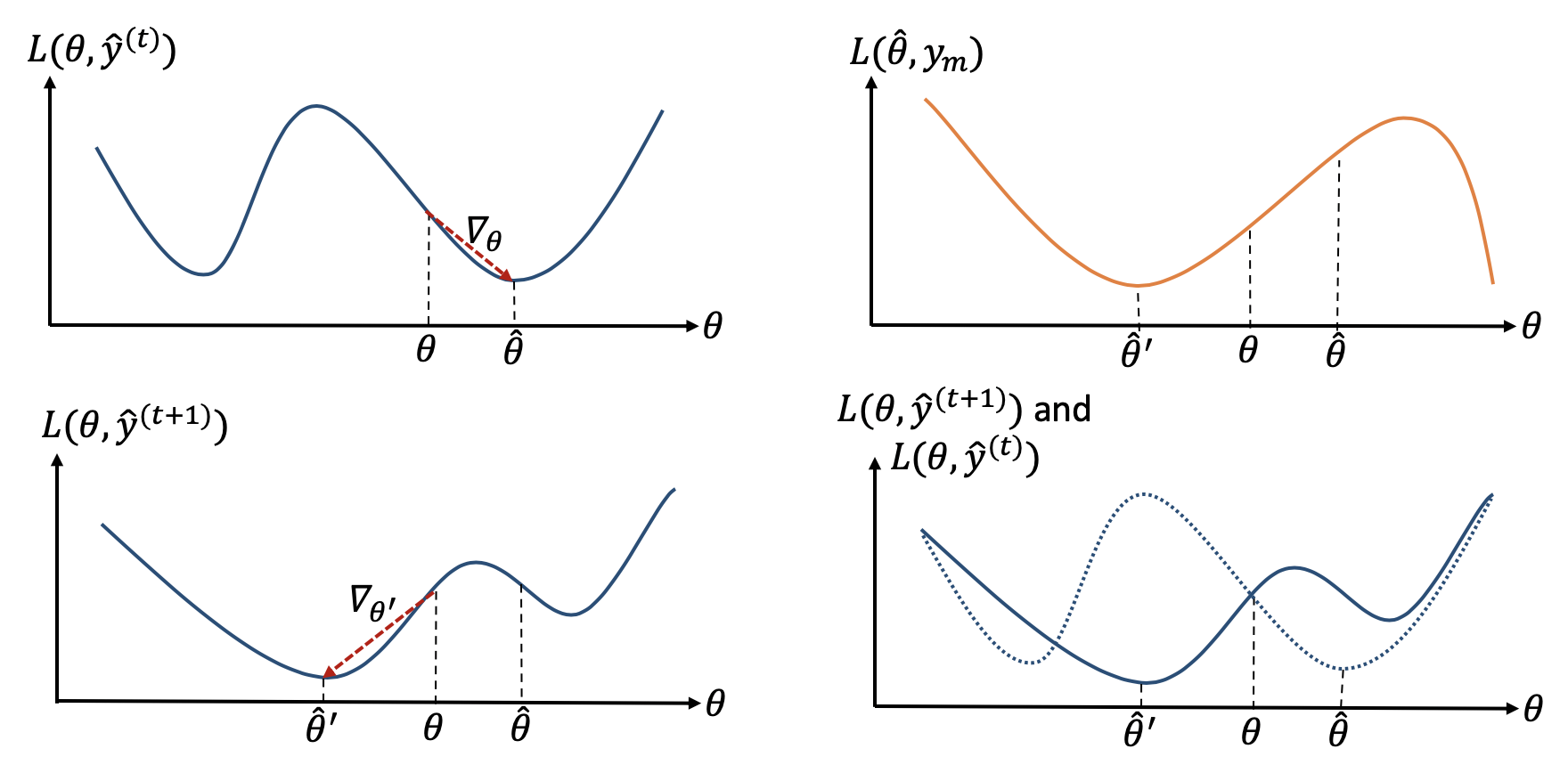}
    \caption{Illustration of the loss function for each step of our algorithm. $L(\theta,\hat{y}^{t})$ is the loss function for training data and predicted labels $\hat{y}^{(t)}=m_{\phi^{(t)}}(v)$, where $v$ represents feature vectors for data $x$. $L(\hat{\theta,}y_m)$ is the loss on meta-data with updated parameter set $\hat{\theta}$. $L(\theta,\hat{y}^{(t+1)})$ is the loss function for training data and updated predicted labels $\hat{y}^{(t+1)}=m_{\phi^{(t+1)}}(v)$. When trained on updated label set $\hat{y}^{(t+1)}$, gradient descent step will result in $\hat{\theta}'$ instead of $\hat{\theta}$, as desired.}
    \label{fig:losses}
\end{figure}

In this formulation $\dfrac{1}{N_b} \sum_{j=1}^{N_b} S_{ij}(\theta,\hat{\theta},\phi^{(t)})$ represents the similarity between the gradient of the $i^{th}$ training sample subjected to model parameters $\theta$ on predicted soft-labels $\hat{y}^{(t)}$ at time step $t$, and the mean gradient computed over the batch of meta-data. As a result, the similarity is maximized when the gradient of $i^{th}$ training sample is consistent with the mean gradient over a batch of meta-data. Therefore, taking a gradient step subjected to $\phi$ means finding the optimal parameter set that would give the best $\hat{y}=m_\phi(g(x))$ such that produced gradients from training data are similar to gradients from meta-data. As a result, the presented meta-objective reshapes the loss function by changing soft-labels so that that resulting gradient updates would lead to model parameters with the minimum loss on meta-data. Therefore, our meta-objective aims to reshape the classification loss function by changing soft-labels to adjust gradient directions for least noise affected learning (\autoref{fig:losses}).

	\section{Experiments} \label{experiments}
We conducted comprehensive experiments on three different datasets. In this section, firstly the description of the used datasets is provided. Secondly, the experimental setup used during the trials is described. Thirdly, results on these datasets are presented and compared to state-of-the-art methods.

\subsection{Datasets}
\subsubsection{\textbf{CIFAR10}}
CIFAR10 \cite{torralba200880} has 60k images for 10 different classes. We separated 5k images for the test-data and another 5k for meta-data. Training data is corrupted with two types of synthetic label noises; uniform noise and feature-dependent noise. For uniform noise, labels are flipped to any other class uniformly with the given error probability. For feature-dependent noise, a pre-trained network is used to map instances to the feature domain. Then, samples that are closest to decision boundaries are flipped to its counter class. This noise is designed to mimic the real-world noise coming from a human annotator. 

\subsubsection{\textbf{Clothing1M}}
Clothing1M is a large-scale dataset with one million images collected from the web \cite{xiao2015learning}. It has images of clothings from 14 classes. Labels are constructed from surrounding texts of images and are estimated to have a noise rate of around 40\%. There exists 50k, 14k and 10k additional verified images for training, validation and test set. We used validation set for meta-data. We did not use 50k clean training samples in any part of the training. 

\ifdefined\FOODN
\subsubsection{\textbf{Food101N}}
Food101N is an image dataset containing 310k images of food recipes belonging to 101 different classes \cite{lee2018cleannet}. It shares the same classes with Food101 dataset but has much more noisy labels, which is estimated to be around 20\%. There exists 53k and 5k verified images for training and test set. We used 15k samples from verified training samples for meta-data.
\fi

\subsubsection{\textbf{WebVision}}
WebVision 1.0 dataset \cite{li2017webvision} consists of 2.4 million images crawled from Flickr website and Google Images search. As a result, it has many real-world noisy labels. It has images from 1000 classes that are same as ImageNet ILSVRC 2012 dataset \cite{russakovsky2015imagenet}. To have a fair comparison with previous works, we used the same setup with \cite{chen2019understanding} as follows. We only used Google subset of data and among these data we picked samples only from the first 50 classes. This subset contains 2.5k verified test samples, from which we used 1k as meta-data and 1.5k as test-data.

\begin{table*}[h]
    \setlength\tabcolsep{2pt}
    \resizebox{\textwidth}{!}{
    \begin{tabular}{l|cccc|cccc}
        \hline
        noise type                               & \multicolumn{4}{c|}{uniform}                                                                          & \multicolumn{4}{c}{feature-dependent}                                                                  \\ \hline 
        noise ratio (\%)                         & 20                      & 40                      & 60                      & 80                      & 20                      & 40                      & 60                      & 80                       \\ \hline 
        Cross Entropy                            & 82.55$\pm$0.80          & 76.31$\pm$0.75          & 65.94$\pm$0.44          & 38.19$\pm$0.81          & 81.75$\pm$0.39          & 71.86$\pm$0.69          & 69.78$\pm$0.71          & 23.18$\pm$0.55           \\        
        SCE\cite{wang2019symmetric}              & 81.36$\pm$2.27          & 78.94$\pm$2.22          & 72.20$\pm$2.24          & 51.47$\pm$1.58          & 74.45$\pm$2.97          & 63.71$\pm$0.58          & fail                    & fail                     \\        
        GCE\cite{zhang2018generalized}           & 84.98$\pm$0.30          & \textbf{81.65$\pm$0.30} & \textbf{74.59$\pm$0.46} & 42.53$\pm$0.24          & 81.43$\pm$0.45          & 72.35$\pm$0.51          & 66.60$\pm$0.43          & fail                     \\        
        Bootstrap \cite{reed2014training}        & 82.76$\pm$0.36          & 76.66$\pm$0.72          & 66.33$\pm$0.30          & 38.35$\pm$1.83          & 81.59$\pm$0.61          & 72.18$\pm$0.86          & 69.50$\pm$0.29          & 23.05$\pm$0.64           \\        
        Forward Loss\cite{patrini2017making}     & 83.24$\pm$0.37          & 79.69$\pm$0.49          & 71.41$\pm$0.80          & 31.53$\pm$2.75          & 77.17$\pm$0.86          & 69.46$\pm$0.68          & 36.98$\pm$0.73          & fail                     \\        
        Joint Opt.\cite{tanaka2018joint}         & 83.64$\pm$0.50          & 78.69$\pm$0.62          & 68.83$\pm$0.22          & 39.59$\pm$0.77          & 81.83$\pm$0.51          & 74.06$\pm$0.57          & 71.74$\pm$0.63          & 44.81$\pm$0.93           \\        
        PENCIL\cite{yi2019probabilistic}         & 83.86$\pm$0.47          & 79.01$\pm$0.62          & 71.53$\pm$0.39          & 46.07$\pm$0.75          & 81.82$\pm$0.41          & 75.18$\pm$0.50          & 69.10$\pm$0.24          & fail                     \\        
        Co-Teaching\cite{han2018co}              & \textbf{85.82$\pm$0.22} & 80.11$\pm$0.41          & 70.02$\pm$0.50          & 39.83$\pm$2.62          & 81.07$\pm$0.25          & 72.73$\pm$0.61          & 68.08$\pm$0.42          & 18.77$\pm$0.07           \\        
        MLNT\cite{junnan2018learning}            & 83.32$\pm$0.45          & 77.59$\pm$0.85          & 67.44$\pm$0.45          & 38.83$\pm$1.76          & 82.07$\pm$0.76          & 73.90$\pm$0.34          & 69.16$\pm$1.13          & 22.85$\pm$0.45           \\        
        Meta-Weight\cite{shu2019meta}            & 83.59$\pm$0.54          & 80.22$\pm$0.16          & 71.22$\pm$0.81          & 45.81$\pm$1.78          & 81.36$\pm$0.54          & 72.52$\pm$0.51          & 67.59$\pm$0.43          & 22.10$\pm$0.69           \\        
        MSLG\cite{algan2020meta}                 & 83.03$\pm$0.44          & 78.28$\pm$1.03          & 71.30$\pm$1.54          & \textbf{52.43$\pm$1.26} & 82.22$\pm$0.57          & 77.62$\pm$0.98          & 73.08$\pm$1.58          & 57.30$\pm$7.40           \\ \hline 
        \textbf{Ours}                            & 83.35$\pm$0.17          & 79.03$\pm$0.26          & 70.60$\pm$0.86          & 50.02$\pm$0.44          & \textbf{83.00$\pm$0.41} & \textbf{80.42$\pm$0.51} & \textbf{78.42$\pm$0.36} & \textbf{76.57$\pm$0.33}  \\ \hline 
    \end{tabular}}
    \caption{Test accuracies for CIFAR10 dataset with varying level of uniform and feature-dependent noise. Results are averaged over 4 runs.}
    \label{tbl:cifar10}
\end{table*} 

\subsection{Implementation Details}
We used SGD optimizer with 0.9 momentum and $10^{-4}$ weight decay for the base classifier in all experiments. Adam optimizer with weight decay of $10^{-4}$ is used for MetaLabelNet. SLP with the size of NUM FEATURES x NUM CLASSES is used for MetaLabelNet architecture. For model selection we used the meta-data and evaluated the final performance on test data. Dataset specific configurations are as follows.

\subsubsection{CIFAR10}
We use an 8-layer convolutional neural network with 6 convolutional layers and 2 fully-connected layers. The batch size is set to 128. $\lambda$ is initialized as $10^{-2}$ and set to $10^{-3}$ and $10^{-4}$ at $40^{th}$ and $80^{th}$ epochs. $\beta$ is set to $10^{-2}$. Total training consists of 120 epochs, in which the first 44 epochs are warm-up. For data augmentation, we used a random vertical and horizontal flip. Moreover, we pad images 4 pixels from each side and random crop 32x32 pixels.

\subsubsection{Clothing1M}
In order to have a fair comparison, we followed the widely used setup of ResNet-50 \cite{he2016deep} architecture pre-trained on ImageNet \cite{deng2009imagenet}. The batch size is set to 32. $\lambda$ is set to $10^{-3}$ for the first 5 epochs and to $10^{-4}$ for the second 5 epochs. $\beta$ is set to $10^{-3}$. Total training consists of 10 epochs, in which the first epoch is warm-up training. All images are resized to 256x256, and then central 224x224 pixels are taken. Additionally, we applied a random horizontal flip.

\ifdefined\FOODN
\subsubsection{Food101N}
For this dataset, we used the same experimental setup that is used for Clothing1M dataset. Only difference is that we keep warm-up training longer, which is 6 epochs.
\fi

\subsubsection{WebVision}
Following the previous works \cite{chen2019understanding}, we used inception-resnet v2 \cite{szegedy2016inception} network architecture with random initialization. The batch size is set to 16. $\lambda$ is set to $10^{-2}$ for the first 12 epochs and to $10^{-3}$ for the rest. $\beta$ is set to $10^{-3}$. Total training consists of 30 epochs, in which the first 14 epochs are warm-up training. All images are resized to 320x320, and then central 299x299 pixels are taken. Furthermore, we applied a random horizontal flip.

\begin{figure}[b]
    \centering
    \includegraphics[width=\columnwidth]{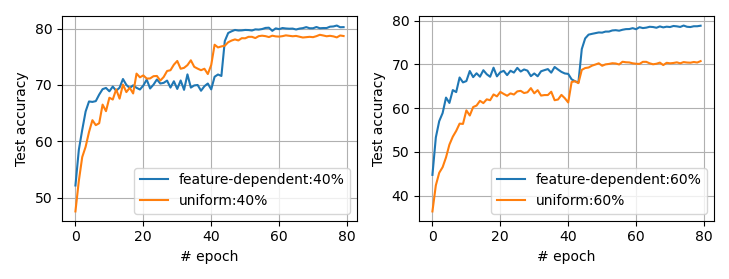}
    \caption{Test accuracies for feature-dependent and uniform noise at noise ratios 40\% and 60\% on CIFAR10 dataset.}
    \label{fig:smvsfd}
\end{figure} 

\subsection{Results on CIFAR10}
We conduct tests with synthetic noise on CIFAR10 dataset and comparison of results to baseline methods is presented in \autoref{tbl:cifar10}. Our algorithm manages to give the best results on all levels of feature-dependent noise and achieves comparable performance for uniform-noise. 

Given training data labels are not used in Algorithm \ref{algo}, but they are used during the warm-up training. As a result, noisy-labels affect initial weights for the base model. In feature-dependent noise, labels are flipped to the class of most resemblance. Therefore, even though it degrades the overall performance, the classifier model still learns useful representations of data. This provides a better initialization for our algorithm. As shown in \autoref{fig:smvsfd}, performance is boosted around 10\% for feature-dependent noise and 5\% for uniform noise. Moreover,  even though initial performance after warm-up training is worse for feature-dependent noise at 40\% noise ratio, it manages a better final performance. From these observations, we can conclude that the proposed algorithm performs better as the noise gets related to the underlying data features. This is an advantage for real-world scenarios since noisy labels are commonly related to data attributes. We showed this further on noisy real-world datasets in the next sections.

\begin{figure}[h]
    \centering
    \includegraphics[width=\columnwidth]{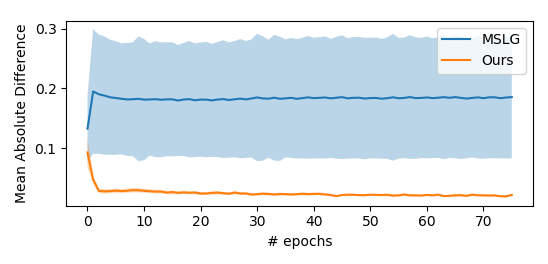}
    \caption{Comparison of the stability of the produced labels with \cite{algan2020meta}. Lines indicates the mean absolute difference between produced labels in consecutive epochs. Shaded region indicates the variance in the differences.}
    \label{fig:mslgvs}
\end{figure}

In \autoref{fig:mslgvs}, we compared the stability of the produced labels \cite{algan2020meta}. This observation is consistent with the higher performance of our proposed algorithm.

\subsection{Results on Clothing1M}
Clothing1M is a widely used benchmarking dataset to evaluate performances of label noise robust learning algorithms. State of the art results from the literature are presented in \autoref{results_clothing1m}, where we managed to outperform all baselines. Our proposed algorithm achieves 78.20\% test accuracy, which is ~3.5\% higher than the closest baseline. 

\begin{table}[H]
    \resizebox{\columnwidth}{!}{
    \begin{tabular}{l|l|l|l}
        \hline
        \multicolumn{1}{c|}{method} & \multicolumn{1}{c|}{acc}  & \multicolumn{1}{c|}{method}  & \multicolumn{1}{c}{acc}    \\ \hline
        MLC\cite{wang2020training}              & 71.10         & Meta-Weight \cite{shu2019meta}       & 73.72              \\
        Joint Opt. \cite{tanaka2018joint}       & 72.23         & NoiseRank \cite{sharma2020noiserank} & 73.77              \\
        MetaCleaner \cite{zhang2019metacleaner} & 72.50         & Anchor points\cite{xia2019anchor}    & 74.18              \\
        SafeGuarded \cite{yao2019safeguarded}   & 73.07         & CleanNet \cite{lee2018cleannet}      & 74.69              \\  
        MLNT \cite{junnan2018learning}          & 73.47         & MSLG\cite{algan2020meta}             & 76.02              \\
        PENCIL \cite{yi2019probabilistic}       & 73.49         & \textbf{Ours}                        & \textbf{78.20}     \\ \hline
    \end{tabular}}
    \caption{Test accuracies on Clothing1M dataset. All results are taken from the corresponding paper.}
    \label{results_clothing1m}
\end{table}

\ifdefined\FOODN
\subsection{Results on Food101N}
Results on Food101N dataset are presented in \autoref{results_food101N}. Some methods fail to succeed since there are excessively large number of classes (101). Therefore, we only presented results which had fair performance.

\begin{table}[H]
    \resizebox{\columnwidth}{!}{
    \begin{tabular}{l|l|l|l}
        \hline
        \multicolumn{1}{c|}{method} & \multicolumn{1}{c|}{acc.} & \multicolumn{1}{c|}{method} & \multicolumn{1}{c}{acc.}\\ \hline
        Joint Opt. \cite{tanaka2018joint}   & 76.12         & PENCIL \cite{yi2019probabilistic}  & 78.26                \\
        Meta-Weight \cite{shu2019meta}      & 76.12         & Co-Teaching \cite{han2018co}       & 78.95                \\  
        Bootstrap \cite{reed2014training}   & 78.03         & \textbf{Ours}                      & \textbf{80.46}       \\ \hline
    \end{tabular}}
    \caption{Test accuracy percentages for Food101N dataset. All values in the table are obtained from our own implementations.}
    \label{results_food101N}
\end{table}
\fi

\subsection{Results on WebVision}
\autoref{results_webvision} shows performances on WebVision dataset. As presented, our algorithm manages to get the best performance on this dataset as well.

\begin{table}[H]
    \centering
    \resizebox{.7\columnwidth}{!}{
    \begin{tabular}{l|l|l}
        \hline
        \multicolumn{1}{c|}{method} & \multicolumn{1}{c|}{Top1}  & \multicolumn{1}{c}{Top5}  \\ \hline
        Forward Loss \cite{patrini2017making}       & 61.12 & 82.68   \\  
        Decoupling \cite{malach2017decoupling}      & 62.54 & 84.74   \\
        D2L \cite{ma2018dimensionality}             & 62.68 & 84.00   \\ 
        MentorNet \cite{jiang2017mentornet}         & 63.00 & 81.40   \\ 
        Co-Teaching \cite{han2018co}                & 64.58 & 85.20   \\ 
        Iterative-CV \cite{chen2019understanding}   & 65.24 & 85.34   \\ \hline
        \textbf{Ours}                               & \textbf{67.10} & \textbf{87.48}   \\ \hline
    \end{tabular}}
    \caption{Test accuracies on WebVision dataset. Baseline results are taken from \cite{chen2019understanding}}
    \label{results_webvision}
\end{table}
	\section{Ablation Study} \label{ablation}
In this chapter, we analyze the effect of individual hyper-parameters on the overall performance.

\begin{figure}[t]
    \centering
    \includegraphics[trim = 0mm 0mm 0mm 90mm, clip, width=\columnwidth]{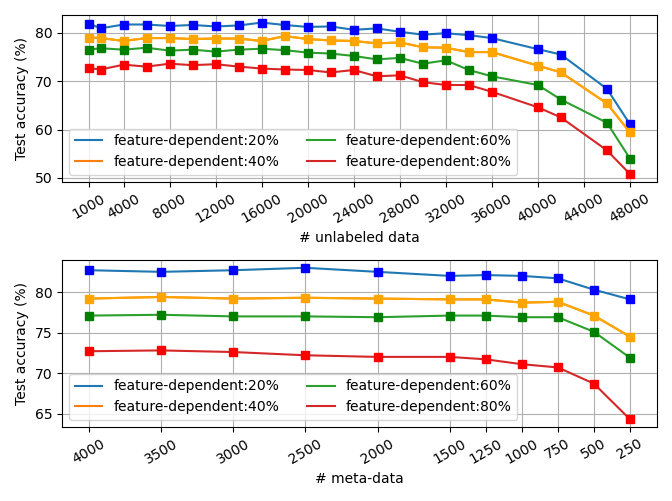}
    \caption{Test accuracies for different levels of feature-dependent noise for varying amount of meta-data on CIFAR10.}
    \label{fig:metadatasizes}
\end{figure} 
\begin{figure}[b]
    \centering
    \includegraphics[width=\columnwidth]{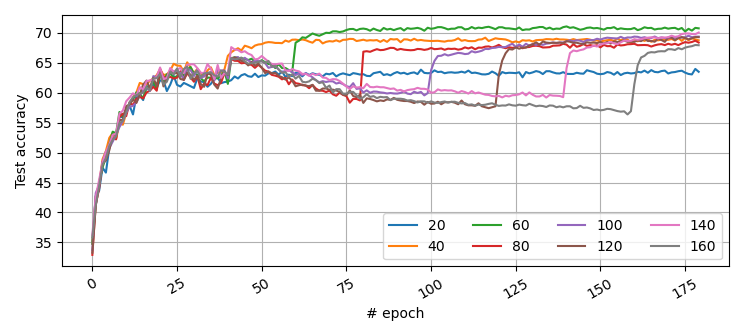}
    \caption{Test accuracies for different duration of warm-up training with 60\% noise on CIFAR10 dataset. Given numbers indicate the number of epochs spend for warm-up training.}
    \label{fig:warmup_duration}
\end{figure}

\subsection{\textbf{Warm-Up Training Duration}}
\autoref{fig:warmup_duration} presents results for different durations of warm-up training. We observed that it is beneficial to train on noisy data as warm-up training until a few epochs later the first learning rate decay. Because, during this duration training is mostly resilient to noise. After that, the base model starts to overfit the noise. As warm-up training gets longer and longer, the base model overfits the noise even more, which leads to poor model initialization for our algorithm. But even then, as soon as we employ the proposed algorithm, test accuracy increases significantly and ends up in a small margin of best performance. On the other hand, if we finish warm-up training early (such as $20^{th}$ epoch), overall performance decreases significantly.

\begin{table}[t]
    \resizebox{\columnwidth}{!}{
    \begin{tabular}{l|l|l|l|l|l|l|l}
        \hline
        \multicolumn{1}{c|}{\# meta-data} & \multicolumn{1}{c|}{0.5k} & \multicolumn{1}{c|}{1k} & \multicolumn{1}{c|}{2k} & \multicolumn{1}{c|}{4k} & \multicolumn{1}{c|}{8k} & \multicolumn{1}{c|}{12k} & \multicolumn{1}{c}{14k} \\ \hline
        Test Accuracy                     & 76.13                     & 77.71                   & 76.95                   & 77.60                   & 77.44                   & 77.61                    & 78.20                   \\ \hline  
    \end{tabular}}
    \caption{Test accuracies for varying number of meta-data for Clothing1M dataset.}
    \label{clothing1mb_metadatasizes}
\end{table}

This is due to less stabilized initial model parameters. Alternatively, if we finish warm-up training exactly at the epoch of learning rate decay ($40^{th}$ epoch), it still gives a sub-optimal performance. Therefore, we advise employing warm-up training until a few epochs later the first learning rate drop and then continue training with Algorithm \ref{algo}.

\subsection{\textbf{$\beta$ Value}}
We used Adam optimizer for the MetaLabelNet and observed that best values of $\beta$ is in between $10^{-2}-10^{-3}$. As the dataset gets simpler (e.g. CIFAR10), $10^{-2}$ results in better performance. On the contrary, $10^{-3}$ provides better performance for more complex datasets (e.g. Clothing1M, WebVision).

\subsection{\textbf{MetaLabelNet Complexity}}
We observed that increasing MetaLabelNet complexity does not contribute to the overall performance. Three different configurations of multi-layer-perceptron (MLP) network with one hidden layer are tested as follows; 1)Sole MLP network 2)Additional batch-normalization layers 3)Additional batch-normalization and dropout layers. None of the configurations manages to get better accuracy than the proposed MetaLabelNet architecture. Our MetaLabelNet is a single, fully-connected layer with the same size as the last fully-connected layer of the base classifier. This is logical because input features to MetaLabelNet are extracted from the last layer before the fully-connected-layer of the base classifier. Therefore, keeping the same configuration as the last layer of the base classifier leads to the easiest interpretation of the extracted features.

\subsection{\textbf{Meta-Data Size}}
\autoref{fig:metadatasizes} and \autoref{clothing1mb_metadatasizes} shows the impact of meta-data size for CIFAR10 and Clothing1M datasets. Around 1k meta-data suffices for top performance in both datasets, which is around 100 samples for each class. It should be noted that Clothing1M is 15 times bigger than CIFAR10 dataset; nonetheless, they require the same number of meta-data. So, our algorithm does not demand an increasing number of meta-data for an increasing number of training data. 

\subsection{\textbf{Unlabeled-Data Size}}
\begin{figure}[t]
    \centering
    \includegraphics[trim = 0mm 90mm 0mm 0mm, clip, width=\columnwidth]{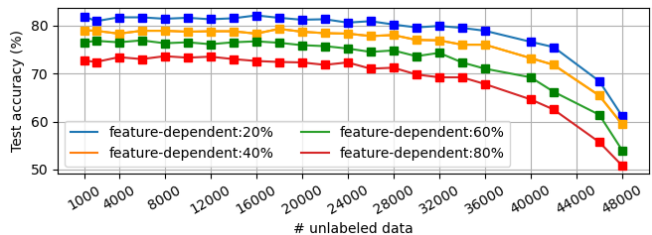}
    \caption{Test accuracies for different levels of feature-dependent noise for varying amount of unlabeled-data on CIFAR10.}
    \label{fig:unlabeled}
\end{figure} 
\begin{table}[t]
    \resizebox{\columnwidth}{!}{
    \begin{tabular}{l|l|l|l|l|l|l}
        \hline
        \multicolumn{1}{c|}{\# unlabeled-data} & \multicolumn{1}{c|}{100k} & \multicolumn{1}{c|}{300k} & \multicolumn{1}{c|}{500k} & \multicolumn{1}{c|}{700k} & \multicolumn{1}{c|}{900k}& \multicolumn{1}{c}{950k} \\ \hline
        Cross Entropy                          & 69.32                     & 68.66                     & 67.87                     & 68.02                     & 68.07                    & 66.88                    \\ \hline
        Ours                                   & 77.90                     & 77.54                     & 77.31                     & 76.52                     & 75.23                    & 71.32                    \\ \hline
    \end{tabular}}
    \caption{Test accuracies for varying number of unlabeled data for Clothing1M dataset.}
    \label{unlabeled_clothing1m}
\end{table}
Labels of the training data are not used in the proposed Algorithm \ref{algo}, but they are required for the warm-up training. Therefore, even though unlabeled-data's size does not directly affect training, it indirectly affects by providing a stable starting point. We removed labels of a certain amount of training data and conducted warm-up training only on the labeled part. Afterward, the whole dataset is used for training with Algorithm \ref{algo}. \autoref{fig:unlabeled} shows the impact of unlabeled-data size for different levels of feature-dependent noise on CIFAR10 dataset. As illustrated, our algorithm can give top accuracy even up to a point where 72\% of the training data (36k images) is unlabeled. Secondly, \autoref{unlabeled_clothing1m} presents results for Clothing1M dataset. For comparison, we trained an additional network with the classical cross entropy method only on labeled data samples and measured the performances. Even in the extreme case of 900k unlabeled data, which is 90\% of the training data, our algorithm manages to achieve 75.23\% accuracy, which is higher than all state-of-the-art baselines.
	\section{Conclusion} \label{conclusion}
In this work, we proposed a meta-learning based label noise robust learning algorithm. Our algorithm is based on the following simple assumption: \textit{optimal model parameters learned with noisy training data should minimize the cross-entropy loss on clean meta-data}. In order to impose this, we employed a learning framework with two steps, namely: meta training step and conventional training step. In the meta training step, we update the soft-label generator network, which is called MetaLabelNet. This step consists of two sub-steps. On the first sub-step, we calculate the classification-loss between model predictions on training data and corresponding soft-labels generated by MetaLabelNet. Then, updated classifier model parameters are determined by taking a stochastic-gradient-descent step on this classification-loss. On the second sub-step, meta-loss is calculated between model predictions on meta-data and corresponding clean labels. Finally, gradients are backpropagated through these two losses for the purpose of updating MetaLabelNet parameters. This meta-learning approach is also called taking gradients over gradients. As a result, \textit{our meta objective seeks for soft-labels such that gradients from classification loss would lead network parameters in the direction of minimizing meta loss}. Then, in the conventional training step, the classifier model is trained on training data and soft-labels generated by updated MetaLabelNet. These steps are repeated for each batch of data. To provide a reliable start point for the mentioned algorithm, we employ a warm-up training on noisy data with classical cross-entropy loss before the learning framework mentioned above. The presented learning framework is highly independent of training data labels, allowing unlabeled data to be used in training. We tested our algorithm on benchmark datasets with synthetic and real-world label noises. Results show the superior performance of the proposed algorithm. For future work, we intend to extend our work beyond the noisy label problem domain and merge our method with self-supervised learning techniques. In such a setup, one can put effort into picking up the most informative samples. Later on, by using these samples as meta-data, our proposed meta-learning framework can be used as a performance booster for the self-supervised learning algorithm at hand.

	\section{Acknowledgements}
	We thank Dr. Erdem Akag\"{u}nd\"{u}z for his valuable feedbacks during the writing of this article.

	\bibliographystyle{IEEEtran}
	\bibliography{references} 

	\begin{IEEEbiography}[{\includegraphics[width=1in,height=1.25in,clip,keepaspectratio,angle=270]{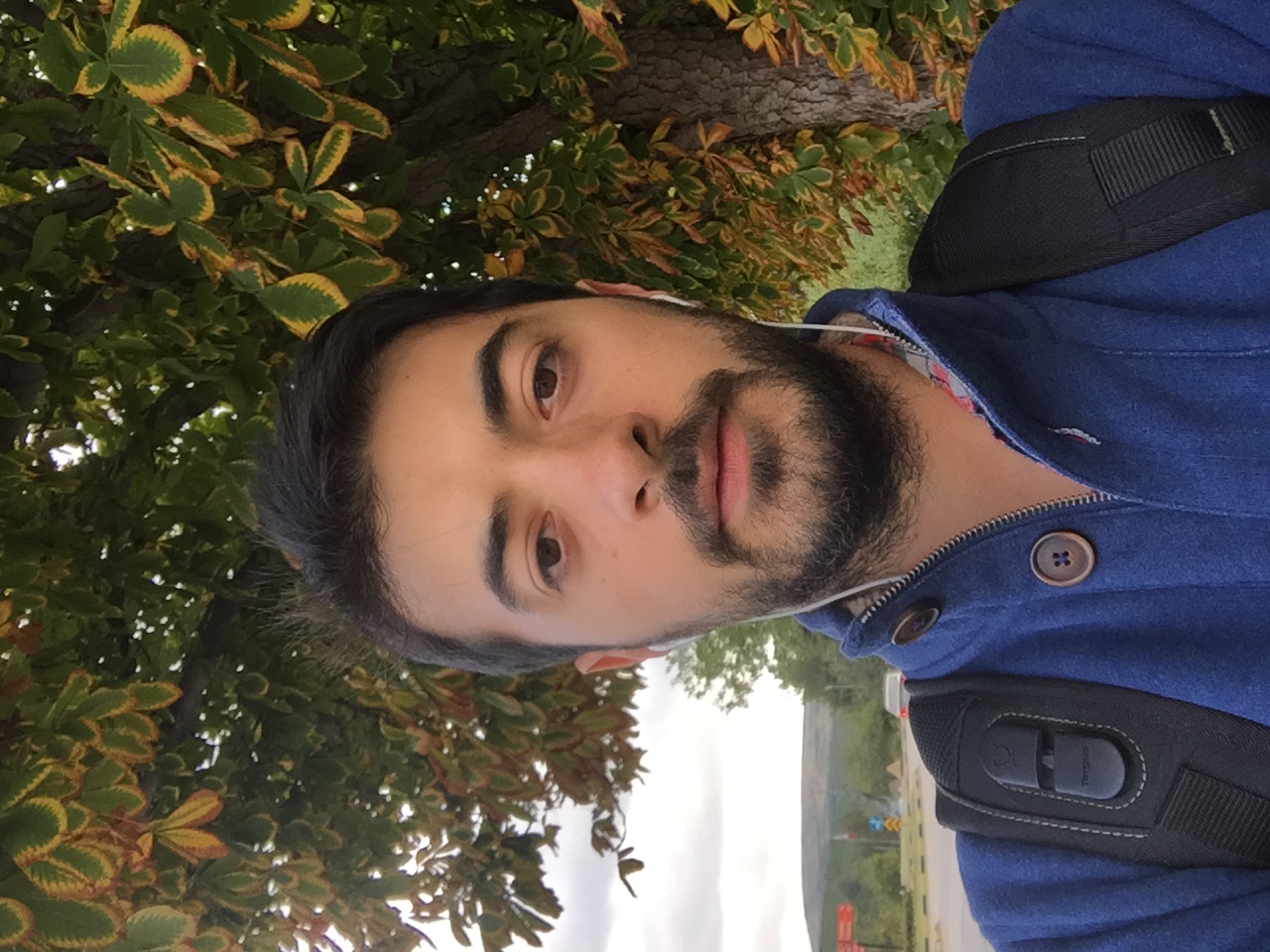}}]{G\"{o}rkem Algan}
		  received his B.Sc. degree in Electrical-Electronics Engineering in 2012, from Middle East Technical University (METU), Turkey. He received his M.Sc. from KTH Royal Institute of Technology, Sweden and Eindhoven University of Technology, Netherlands with double degree in 2014. He is currently a Ph.D. candidate at the Electrical-Electronics Engineering, METU. His current research interests include deep learning in the presence of noisy labels.
	\end{IEEEbiography}

	\begin{IEEEbiography}[{\includegraphics[width=1in,height=1.25in,clip,keepaspectratio]{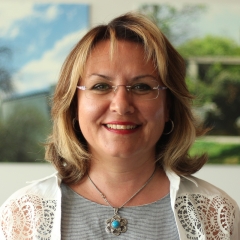}}]{Ilkay Ulusoy}
		was born in Ankara, Turkey, in 1972. She received the B.Sc. degree from the Electrical and Electronics Engineering Department, Middle East Technical University (METU), Ankara, in 1994, the M.Sc. degree from The Ohio State University, Columbus, OH, USA, in 1996, and the Ph.D. degree from METU, in 2003. She did research at the Computer Science Department of the University of York, York, U.K., and Microsoft Research Cambridge, U.K. She has been a faculty member in the Department of Electrical and Electronics Engineering, METU, since 2003. Her main research interests are computer vision, pattern recognition, and probabilistic graphical models.	
	\end{IEEEbiography}

\end{document}